# Vision-Ultrasound Robotic System based on Deep Learning for Gas and Arc Hazard Detection in Manufacturing

Jin-Hee Lee[1], Dahyun Nam[2], Robin Inho Kee[3], YoungKey Kim[4], Seok-Jun Buu*[1]

[1] Dept. of Computer Engineering, Gyeongsang National University, Jinju 52828, South Korea

[2] Dept. of Mechanical Engineering, Seoul National University, Seoul 08826, South Korea

[3] Dept. of Mechanical Engineering, University of Michigan, Ann Arbor, MI 48109, USA

[4] SM Instruments Inc., Daejeon 34109, South Korea

gsm04548@gnu.ac.kr, diana_nam@snu.ac.kr, inhokee@umich.edu, youngkey@smins.co.kr, sj.buu@gnu.ac.kr*

**Abstract**

Gas leaks and arc discharges present significant risks in industrial environments, requiring robust detection systems to ensure safety and operational efficiency. Inspired by human protocols that combine visual identification with acoustic verification, this study proposes a deep learning-based robotic system for autonomously detecting and classifying gas leaks and arc discharges in manufacturing settings. The system is designed to execute all experimental tasks (A, B, C, D) entirely onboard the robot without external computation, demonstrating its capability for fully autonomous operation. Utilizing a 112-channel acoustic camera operating at a 96 kHz sampling rate to capture ultrasonic frequencies, the system processes real-world datasets recorded in diverse industrial scenarios. These datasets include multiple gas leak configurations (e.g., pinhole, open end) and partial discharge types (Corona, Surface, Floating) under varying environmental noise conditions. The proposed system integrates YOLOv5 for visual detection and a beamforming-enhanced acoustic analysis pipeline. Signals are transformed using Short-Time Fourier Transform (STFT) and refined through Gamma Correction, enabling robust feature extraction. An Inception-inspired Convolutional Neural Network further classifies hazards, achieving an unprecedented 99% gas leak detection accuracy. The system not only detects individual hazard sources but also enhances classification reliability by fusing multi-modal data from both vision and acoustic sensors. When tested in reverberation and noise-augmented environments, the system outperformed conventional models by up to 44%p, with experimental tasks meticulously designed to ensure fairness and reproducibility. Additionally, the system is optimized for real-time deployment, maintaining an inference time of 2.1 seconds on a mobile robotic platform. By emulating human-like inspection protocols and integrating vision with acoustic modalities, this study presents an effective solution for industrial automation, significantly improving safety and operational reliability.

**Keywords:** Gas leak detection, Arc discharge classification, Multimodal deep learning, Ultrasonic beamforming, Industrial hazard detection

# 1. Introduction

Flammable gas leaks and arc discharges are frequently identified as critical safety and operational hazards in industrial settings. When these two phenomena occur simultaneously, there is a heightened probability of explosive incidents, which can lead to severe facility damage, equipment failure, and human casualties [1, 2]. The resultant downtime of machinery further escalates economic losses and contributes to environmental pollution [3-8]. Conventional approaches to detecting and classifying gas leaks and arc discharges typically rely on direct onsite inspections or fixed sensors monitored by operators. However, not all leak or discharge signals manifest in the audible frequency range; many appear in the ultrasonic domain, impeding timely detection and response. Moreover, when sensors are placed too far from the actual leak or discharge location, they may fail to capture critical events. These limitations degrade detection accuracy, reduce efficiency, and ultimately compromise system reliability [9-12].

Previous studies attempted to automate gas leak and discharge detection by combining plant-floor sensor data with deep learning models[13, 14], but they did not fully account for the diverse constraints of real industrial environments, such as intense noise, reverberation, and limited installation space[15]. Consequently, accurately capturing and classifying ultrasonic target signals in factory settings remains a formidable challenge[16]. To address this issue, the present work proposes BATCAM MX, a multichannel deep learning–based autonomous robotic system designed to autonomously navigate to an optimal vantage point for signal acquisition and process both visual and acoustic data by emulating a human-inspired protocol (Human Protocol).

As illustrated in Figure 1-a, a navigation algorithm is composed of line-following and 2D tag-based localization QR code, enabling efficient exploration of the factory floor. Meanwhile, the BATCAM FX acoustic camera, equipped with 112 microphones and RGB camera, captures ultrasonic signals and video streams at frequencies up to 48 kHz and 25FPS each. As shown in Figure 1-b, potential leak or discharge locations identified by the YOLOv5 (You Only Look Once) model are subsequently targeted and amplified through a beamforming algorithm [17, 18], thereby minimizing extraneous interference. The extracted target signals are then meticulously analyzed using STFT and Gamma Correction, after which the resulting spectrograms are segmented via a sliding window approach and provided as inputs to an Inception-style Convolutional Neural Network (CNN). By learning features at multiple time–frequency scales[19-21], the Inception architecture achieves high classification accuracy even for high-dimensional or sub-audible signals. This multi-process of robot navigation, target detection, signal detection and deep learning classification is managed upon Robotic Operating System (ROS) using different communication protocols including CAN, WebSocket, RTSP enabling autonomous mobile robotic platform that can detect gas and arc hazard with an inference time under 2.1s.

The system proposed in this study has been validated under four representative scenarios (Task A, B, C, D) likely to arise in real manufacturing plants, confirming robust detection performance in environments characterized by strong reverberation and intense industrial noise. Notably, this approach achieves up to a 44 percentage-point improvement over existing techniques, with gas leak and arc discharge detection rates reaching as high as 99%. These results are attributed to the fusion of auditory and visual information, which emulates human inspection protocols, along with the precise classification of challenging ultrasonic signals through beamforming and advanced deep learning techniques.

The contributions of this study are as follows:

- Autonomous Multichannel Deep Learning System for Ultrasonic Inspection: Proposes an integrated platform that emulates human-like visual and auditory protocols (Human Protocol) to detect flammable gas leaks and arc discharges in industrial settings.

- Beamforming–YOLO Pipeline for Robust Ultrasonic Signal Enhancement: Selectively amplifies signals in noisy, reverberant environments using beamforming, while YOLOv5 pinpoints suspicious regions for focused analysis.

- Inception-Based CNN on Sliding-Window STFT Spectrograms: Extracts multi-scale time–frequency features to achieve state-of-the-art performance in the ultrasonic domain.

- Validation Across Four Realistic Scenarios: Demonstrates the practical applicability and problem-solving efficacy of the proposed system by autonomously executing all experimental tasks (A, B, C, D) entirely onboard the robot. The system effectively addresses reverberation and severe industrial noise, ensuring robust performance in real-world conditions.

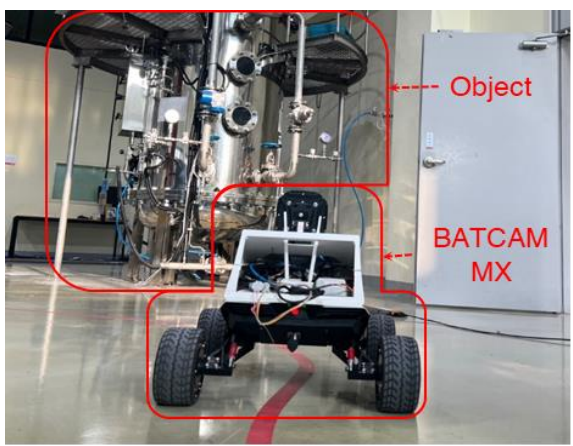
(a) Gas Leakage Detection using Yolov5 on the BATCAM MX

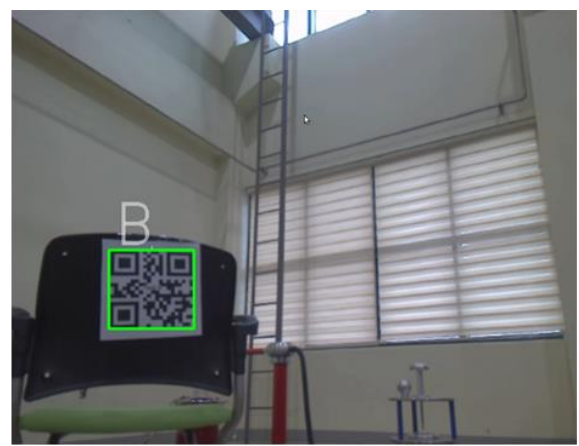
(b) QR Code Detection using BATCAM FX on the BATCAM MX

**Figure 1. An autonomous deep learning-based system (BATCAM MX) performing tasks for gas leakage and discharge detection in a factory environment.**

## 2. Related Works

Addressing gas leaks, power system faults, and partial discharge problems in modern industrial settings has increasingly involved the deployment of diverse sensors and AI-driven methods. As summarized in Table 1, research on automated methane leak detection using infrared imaging has employed frame subtraction or z-score normalization as preprocessing steps for CNN- and LSTM-based models, achieving accuracies of up to 99% and F1-scores of 0.99[22, 23]. By contrast, approaches leveraging visual video signals [24] process camera data from an Autonomous Underwater Vehicle via YOLOv4 or Faster RCNN to detect gas leak plumes in real time, even in complex underwater environments. In the realm of gas leak detection using acoustic signals, experiments conducted on a small-scale gas turbine setup [25] apply CWT (Continuous Wavelet Transform) preprocessing and feed the resulting spectrograms to a ResNet-50 (CNN), attaining a peak accuracy of 90%. Additionally, an indoor, SLAM-based leak detection system employing an autonomous robot [26] demonstrates precise tracking accuracy within 1 m. Furthermore, research integrating GNNs (GNN with Variational Bayesian Inference) into pressure-sensor data [27] achieves strong performance (AUC 0.9484, PAC 0.8) in an urban pipeline network. Power system fault detection and partial discharge diagnosis have also garnered considerable attention. One study [28] that extracts time-series features from current signals and applies ensemble machine learning (e.g., Bagging, Boosting, Stacking) maintains accuracies of up to 99.95% under varying load conditions (5–20 A). Another investigation [29] classifies CWT-transformed current waveforms using a channel-wise thresholding approach (DRSN-CW), reporting an average accuracy of 97.72%. For electromagnetic signals observed in medium-voltage (MV) power lines, a 1D-CNN and autoencoder ensemble was proposed [30], reaching over 91% accuracy and an MCC of 0.645 on both validated and unvalidated real-world datasets. Meanwhile, partial discharge detection systems have commonly employed acoustic or ultrasonic data. In particular, CNN models augmented with Bayesian optimization [31] have achieved over 94% accuracy using STFT preprocessing of ultrasonic signals from transformers and switchgear, contributing to more effective early fault prediction in industrial machinery. Lastly, in high-noise substation environments, a wavelet-based denoising approach has been proposed for partial discharge detection [32]. After converting time-domain signals into the frequency domain, the method applies the DAMAS2 algorithm to a microphone array. Taken together, research in gas-leak detection and power-system fault diagnosis spans a wide spectrum of sensor modalities (infrared, acoustic, electromagnetic, pressure), preprocessing techniques (STFT, CWT, frame subtraction, normalization), and model architectures (CNN, LSTM, GNN, ensemble ML). Nonetheless, practical implementation in noisy, large-scale industrial facilities requires enhanced noise-reduction schemes, multi-sensor data fusion, and integration with autonomous robotic systems for real-time operation. Building on these developments, this study presents an automated robot capable of detecting both gas leaks and discharge events in a complex factory environment by leveraging ultrasonic and visual data. The key contribution lies in combining beamforming with an Inception-style CNN to effectively suppress noise, while also employing robot control algorithm for enhanced in-field applicability. This strategy differentiates itself from prior IR-based or simple acoustic methods by offering greater scalability, as well as robust performance and real-time detection capabilities in noisy conditions.

**Table 1. Overview of related works on signal classification using deep learning models for gas leakage and discharge detection.**

| Objectives | Domain | Representation (Preprocessing) | Method | Environment | Dataset | Performance | Year |
|---|---|---|---|---|---|---|---|
| IR-based Methane detection | Infrared imaging [22] | Background subtraction, image normalization | GasNet (CNN) | Controlled gas leak test facility (METEC) | GasVid with ~1M frames | Accuracy up to 99%, ≥95% overall | 2020 |
| Gas leak detection | Infrared video [23] | Frame subtraction, z-score normalization | CNN + LSTM network | Realistic gas leakage scenarios | Large-scale IR video dataset with varying conditions | Accuracy 98%, F1 0.99, MCC 0.98 | 2022 |
| | Visual signals (video) [24] | Image annotation (LabelImg) | YOLOV4, Faster RCNN | Autonomous underwater vehicle (robotic fish) | 8622 gas leak plume images | YOLOV4: 71ms, Faster RCNN: 3057ms | 2022 |
| | Acoustic signals [25] | Continuous Wavelet Transform (CWT) | ResNet-50 (CNN) | Small-scale gas turbine | 2,578 AE samples across 4 conditions | Accuracy > 80%, up to 90% | 2023 |
| | Indoor environments [26] | Mapping and localization using SLAM | Autonomous mobile robot with LIDAR and PID | Indoor university environment | Simulated acetone gas leak | Detection accuracy within 1m | 2024 |
| | Pressure signals [27] | Normalization, time-series segmentation | VB_GAnomaly (Attention-based GNN with Variational Bayesian Inference) | Urban gas pipeline network | Experimental pipeline network with 5 leak locations | AUC 0.9484, Positioning Accuracy (PAc) 0.8 | 2023 |
| fault detection in power systems | Current signals [28] | Time-domain feature extraction, feature importance | Ensemble ML (Bagging, Boosting, Stacking) | DC power systems with various loads | 5-20A current data in various load conditions | Accuracy up to 99.95% | 2020 |
| | Current signals [29] | Continuous Wavelet Transform (CWT) | DRSN-CW with channel-wise thresholding | Indoor power distribution system | Enhanced data from 272 samples to 4080 | Accuracy up to 98.92%, avg. 97.72% | 2022 |
| Partial discharge detection | Electromagnetic signals [30] | Denoising, max-pooling, mean subtraction | Ensemble stacking (1D-CNN and Autoencoders) | Overhead medium voltage power lines | Real-world dataset with validated, non-validated series | Accuracy ≥91%, MCC 0.645 | 2023 |
| | Acoustic signals [31] | Short-Time Fourier Transform (STFT) | Optimized CNN with Bayesian tuning | Field equipment (transformers, switchgears) | Ultrasound data with 5 classes | Accuracy ≥94% | 2024 |
| | Acoustic signals [32] | Wavelet-based denoising, time-domain to frequency-domain conversion | Wavelet-based denoising, time-domain to frequency-domain conversion | DAMAS2 algorithm with microphone array | High-noise substation environment | Qualitative evaluation | 2024 |

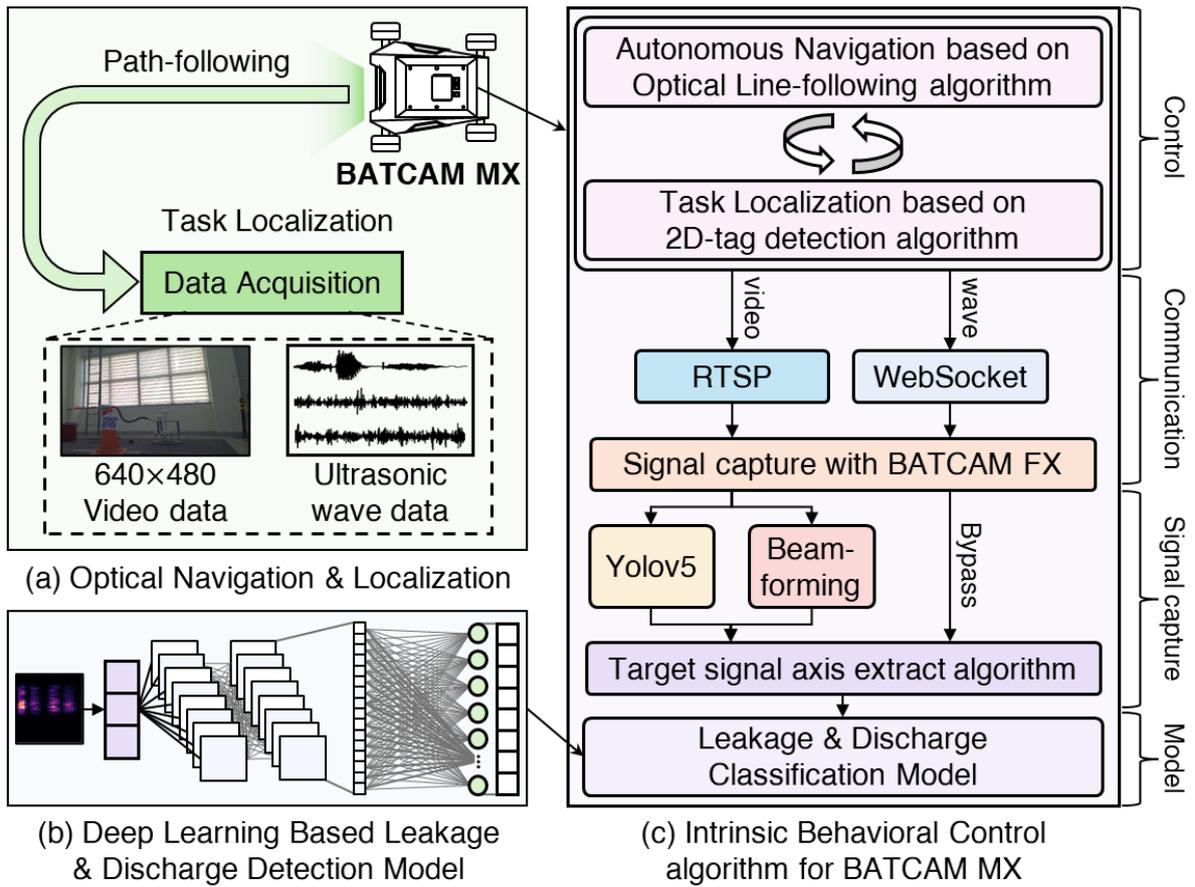

Figure 2. BATCAM MX system overview: overall architecture of the line-tracing-based autonomous vehicle integrated with a deep learning system. QR-based autonomous control and beamforming for deep learning-based signal detection and classification.

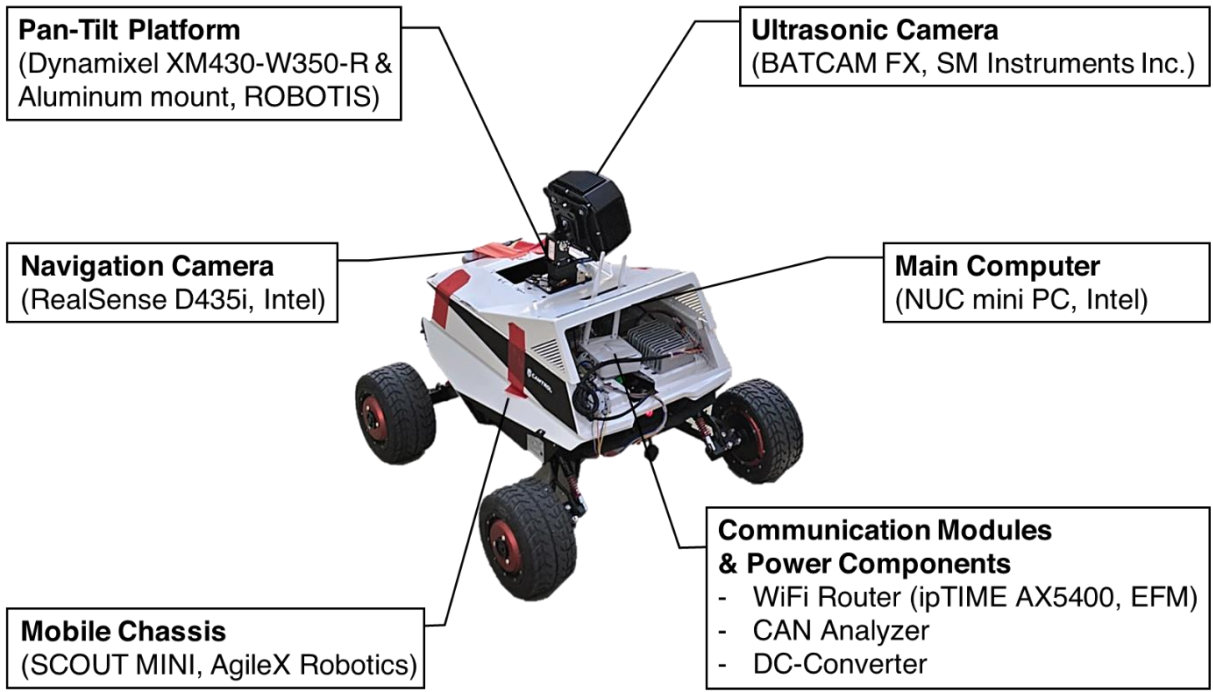

Figure 3. Hardware Overview of the BATCAM MX: The robot is built on a mobile chassis and equipped with computer, sensors, actuators, communication modules, and power components.

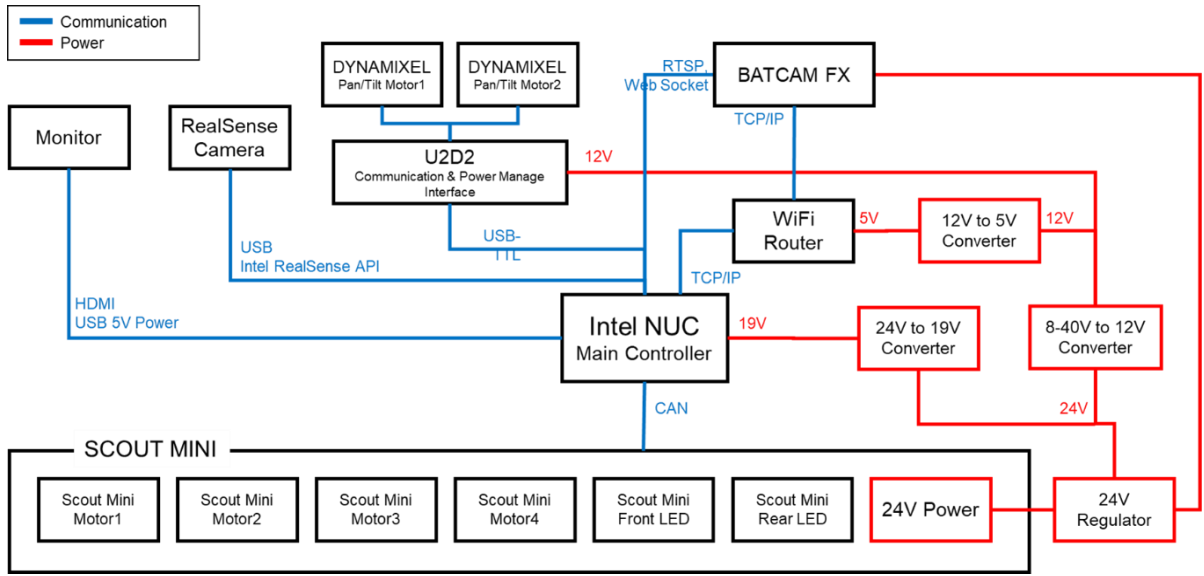

**Figure 4.** Communication diagram of the main components of BATCAM MX: All control and communication are managed by the Intel NUC. The BATCAM FX communicates with the main board via RTSP, WebSocket, and REST API, while motor control is handled through CAN communication.

## 3. The Proposed Method

This study proposes a robotic system: BATCAM MX, combined with a deep learning–based leak and arc discharge detection scheme to address the challenges outlined in the Introduction and Abstract. By implementing autonomously guided optical navigation based on line-following and tag-based positioning, the robot can accurately reach designated targets in a manufacturing environment. Also, by using pan-tilt system with object detection module, the robot can accurately align the acoustic camera's listening point to the target and acquire ultrasonic data. This data is subsequently enhanced via beamforming to more precisely capture target signals. Finally, an Inception-style CNN classifies whether a leak or an arc discharge has occurred, enabling high detection accuracy and real-time performance even in noisy factory environments. Notably, the Inception-style CNN serves as the focal point of the proposed system by extracting features across multiple time–frequency scales, thereby achieving superior classification performance compared to conventional methods. This section presents the hardware and software configuration of the BATCAM MX (Section 3.1), the navigation and localization algorithm to autonomously drive the robot (Section 3.2), the target signal detection and enhancement process (Section 3.3), and deep learning-based leak and discharge detection model to determine any gas leak or discharge (Section 3.4).

### 3.1. BATCAM MX: Robotic System for Autonomous Leak and Discharge Detection

BATCAM MX integrates advanced hardware and software components into a single robotic platform to achieve precise leak and discharge detection in industrial environments. As illustrated in figure 3, the system's hardware is composed of the mobile robot platform, custom power circuit for each sensor, BATCAM FX device, pan-tilt actuator (Dynamixel XM430-W350-R, ROBOTIS) to align the acoustic camera to the target, and a camera in front of the robot for navigation. For the software, as illustrated in Figures 2 and 4, the system architecture combines a line-following-based navigation algorithm, 2D tag-based localization, and beamforming techniques for robust signal detection and classification. Single RGB camera (RealSense Depth Camera D435i, Intel) was utilized to identify the path drawn in the test area and detect the marked QR tags during the path which represents each task target. A proportional–integral–derivative (PID) controller ensures stable line-following and 2D tag detection-based navigation showed robust localization even in dynamic factory settings. When the robot reaches a certain task point indicated by the 2D tag on the track, the robot can perform a detailed and accurate scan of the target location via object detection-based pan-tilt system. During this operation, the BATCAM FX captures acoustic and visual data, which is transmitted to the robot's server in real time via Wi-Fi communication. The robot receives the data through WebSocket and RTSP protocol and processes the data using beamforming algorithms to suppress background noise, enhance target signals. The whole process of autonomously navigating through the track, localizing the task point, scanning and acquiring visual and wave data, and performing leak and discharge detection is systematically managed on the robot's main computer.

### 3.2. Autonomously Guided Robot Navigation Algorithm

The robot's autonomously guided navigation algorithm applied in this research uses single RGB camera to follow the lane and identify the 2D tags installed on the field. Real-time video was streamed at the resolution of 640×480 and the frames were captured then converted to the HSV color space. To identify the QR code, binary thresholding was used for effective tag detection and to identify the target lane, specific color ranges are filtered out. When the target lane is detected, the center point of the line is calculated and used as current input of the PID controller, while the frame center value is used as the reference point. The output from the controller is then used to calculate the control input to the robot's driving controller as target linear and angular velocity by ROS message. When a QR code is identified, the robot stops at the relevant coordinates to perform the leak or discharge detection task. This sequence ensures concurrent autonomous navigation and signal detection in a complex factory environment.

### 3.3. Target Component Detection and Signal Enhancement

This section proposes a method that integrates YOLOv5-based object detection and beamforming to reliably detect target components and enhance the signals. The YOLOv5 model was trained using total 4,800 images collected from the RGB camera of the BATCAM FX device, composed of 1,200 raw image frames and 3,600 augmented images. Data augmentation was done using exposure adjustment, Gaussian noise injection, and filtering considering the different light conditions in the manufacturing line or video frame quality of the robot. The model was then deployed on the robot. RGB video stream was sent from BATCAM FX to the robot's computer (NUC PC, Intel) via websocket and the YOLOv5 model is operated on the video 640×480×3 frame at the framerate of 5 FPS. As the output, the model analyzes the visual information to accurately identify potential risk components (e.g., Flush Rings, Gas Regulators, and Flanges) on the robot's view. When the bounding boxes and class probabilities of detected objects exceed a predefined threshold, the center points were calculated and compressed with the class information. After the process, the compressed information was used to align the BATCAM FX's listening point to every potential risk component detected in the frame.

Once the BATCAM MX receives the ultrasonic signals captured at each listening point (the detected potential risk components), the robot applies a beamforming algorithm to focus on gathering and analyzing ultrasonic signals from the identified steering direction. The ultrasonic signals, captured by the 112-channel BATCAM FX system, are processed to amplify signals from a specific direction while suppressing or canceling noise from other directions. The time-domain beamforming equations are as follows:

$$\tau_m(\theta) = \frac{p_m \cdot u(\theta)}{c},$$

$$y(t;\theta) = \sum_{m=1}^{M} \omega_m x_m(t - \tau_m(\theta))$$

(1)

where $\theta$ represents the steering direction identified by the YOLOv5 model, $p_m$ is the position vector of the $m$-th sensor (centered at the microphones of the BATCAM FX), and $u(\theta)$ is the steering vector in the direction of $\theta$. $c$ denotes the speed of sound, $\tau_m(\theta)$ is the time delay experienced by the $m$-th sensor when receiving a wave from direction $\theta$, and $\omega_m$ is the weight assigned to the $m$-th sensor. $x_m(t)$ is the raw signal received by the $m$-th sensor, and $y(t;\theta)$ is the beamformed signal emphasizing the components arriving from direction $\theta$. Beamforming significantly improves the signal-to-noise ratio (SNR) by selectively amplifying ultrasonic signals arriving from the target direction while simultaneously suppressing or attenuating unwanted noise originating from other directions. This process enhances the clarity and reliability of the received signal, making it more suitable for downstream processing. The resulting 1D signal $y(t;\theta)$ is processed further to classify gas leaks or arc discharges.

**3.4. Deep Learning-Based Leak and Discharge Detection Model**

Ultrasonic signals observed in factory environments typically occupy higher frequency ranges than standard audio signals, making accurate classification challenging due to noise and reverberation. Conventional CNN-based methods often fail to achieve high performance in such conditions. To address these limitations, this study adopts an Inception-style CNN, as shown in Figure 5, designed to simultaneously capture multi-scale time-frequency features. The Inception-style CNN effectively represents complex ultrasonic patterns by employing multiple parallel convolution kernels and stacked Inception blocks for hierarchical feature extraction.

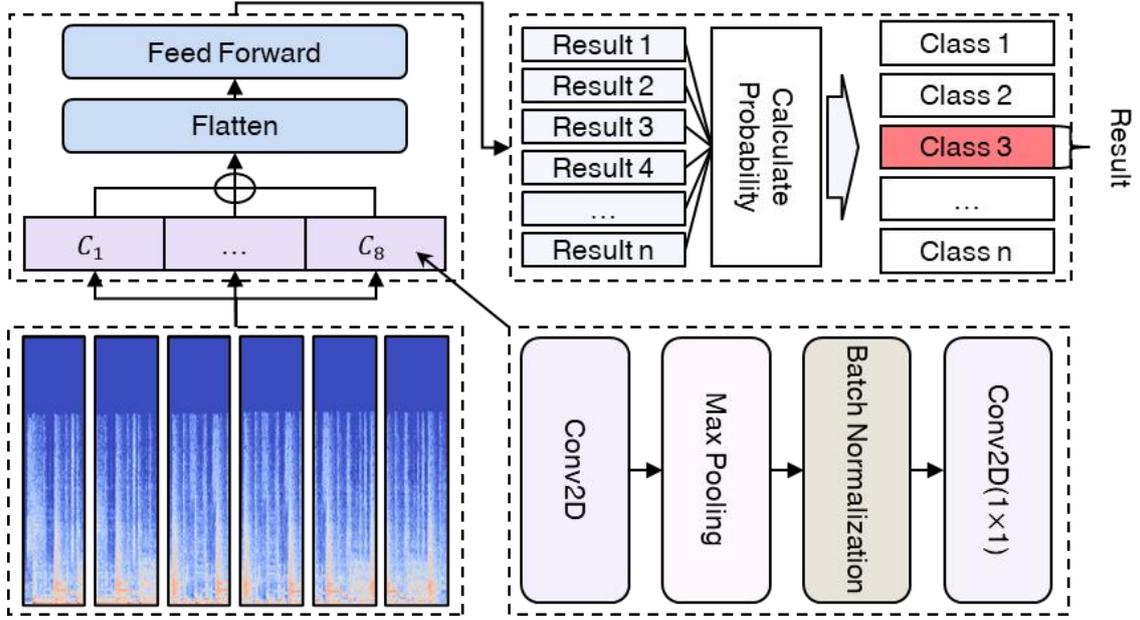

**Figure 5.** Spectrogram data segmented by the sliding window method is processed using the Inception-style CNN. The resulting class probabilities are averaged across all labels to determine the final classification.

### 3.4.1 Input Data Preprocessing

The target signals for detection are ultrasonic data collected from the 112 channels of the BATCAM FX system. These signals are initially segmented into 0.04-second intervals and transformed into spectrograms using the Short-Time Fourier Transform (STFT) to preserve both time and frequency information. The STFT is defined as follows:

$$STFT\{x(t)\}(n, \omega) = \sum_{\tau=-\infty}^{\infty} x(\tau) w(\tau = nH) e^{-j\omega\tau} \qquad (2)$$

where $x(\tau)$ represents the raw signal (i.e., the beamformed signal $y(t;\theta)$), $w(\cdot)$ is the window function (Hamming window in this study), $H$ denotes the frame shift, and $\omega$ is the frequency axis variable. The resulting spectrogram is a 2D time-frequency representation. To optimize computational efficiency and reduce inference time, the spectrogram is divided into smaller segments using a sliding-window sampling technique, expressed as follows:

$$X_i(k) = X(k + i\Delta k), \qquad i = 0, 1, 2, \dots \text{ and } k \in [0, K] \qquad (3)$$

where $X(\cdot)$ represents the spectrogram obtained via STFT, $i$ is the sliding window index, $\Delta k$ is the stride of the sliding window, and $K$ represents the maximum frequency range processed in one operation. To further refine the data, Gamma Correction is applied to suppress residual noise and enhance meaningful features in the target signals. The Gamma Correction is defined as follows:

$$I_{output} = I_{input}^{\gamma} \qquad (4)$$

where *input* represents the input frame, *output* is the gamma-corrected frame, and $\gamma$ is the gamma value that determines the intensity of the correction. When $\gamma > 1$, weaker signals in the spectrogram are attenuated further. This study experimentally determined an optimal $\gamma$ value to suppress non-target signals, such as background noise, while preserving critical gas leak and discharge signals. The gamma-corrected spectrogram segments are then used as inputs to the deep learning model, ensuring task-relevant features are emphasized while irrelevant information is minimized.

### 3.4.2 Overview of the Inception Block

The core component of the proposed model is the Inception Block, which applies Conv2D filters of varying sizes in parallel to extract features across multiple time-frequency scales. This approach is particularly effective in industrial ultrasonic environments characterized by high noise and reverberation. Within each Inception Block, the input tensor $X \in \mathbb{R}^{H \times W \times C_{in}}$ is processed through multiple parallel paths, with each path employing different kernel sizes. For each path $i$, the convolution operation is expressed as:

$$H^{(i)} = f_{BN}(f_{pool}(\sigma(X * W^{(i)} + b^{(i)}))), i = 1, \dots, N \tag{5}$$

where $W^{(i)}$ and $b^{(i)}$ are the kernel weights and biases for path $i$, $\sigma(\cdot)$ is the ReLU activation function, $f_{pool}$ denotes Max Pooling, and $f_{BN}$ represents Batch Normalization. These operations allow each path to extract features at different scales. To unify the output dimensions, the feature maps $H^{(i)}$ are processed through a 1×1 convolutional layer (projection). The resulting outputs are concatenated along the channel dimensions to produce the final output of the Inception Block:

$$H_{proj}^{(i)} = \sigma\left(\left(H^{(i)} * W_{1\times1}^{(i)}\right) + b_{1\times1}^{(i)}\right),$$
$$H_{inception} = Concat(H_{proj}^{(1)} + H_{proj}^{(2)} + \cdots, H_{proj}^{(N)}) \tag{6}$$

This design enables the Inception Block to combine local and global patterns extracted from the ultrasonic spectrum, effectively capturing fine-grained and broad-scale features. The resulting features are flattened into a 1D vector and passed through Multi-Layer Perceptron (MLP) layers for classification.

### 3.4.3 Overview of the Inception Block

This study employs categorical crossentropy as the loss function, commonly used for multi-class classification tasks. The categorical crossentropy loss is mathematically expressed as follows:

$$\mathcal{L}(y, \hat{y}) = -\sum_{i=1}^{m} \hat{y}_j \, log(\hat{y}_j) \tag{7}$$

where $y = (y_1, \dots, y_n)$ represents the ground truth labels, $\hat{y}$ denotes the predicted probability distribution by the model, and mmm is the total number of samples in the dataset. The optimizer parameters, such as the learning rate and batch size, were determined experimentally through preliminary evaluations to ensure robust training. The proposed model's training procedure consists of the following key steps:

1. STFT and Gamma Correction: The raw ultrasonic signals are first converted into spectrograms using STFT. Gamma correction is then applied to emphasize the target features while suppressing noise, thus enhancing the quality of the input data.

2. Inception Block Feature Extraction: Multi-scale features are extracted from the spectrograms using the Inception Blocks, effectively capturing complex ultrasonic patterns across varying time-frequency scales.

3. MLP Classification: The features extracted by the Inception Blocks are flattened into a one-dimensional vector and passed through Multi-Layer Perceptron (MLP) layers to classify events into gas leaks or arc discharges.

To mitigate overfitting, an Early Stopping technique is employed, which terminates the training process if the validation accuracy does not improve after a specified number of epochs. By integrating STFT-Gamma Correction preprocessing with the Inception CNN architecture, the proposed method achieves superior accuracy and stable inference times under complex factory conditions compared to conventional deep learning-based approaches.

### 3.4.4 Post-Processing of Inference Results

The classification results produced by the deep learning model are computed independently for each sliding window segment of the spectrogram. This approach may fail to fully capture the continuity of the overall signal. As illustrated in Figure 5, the BATCAM FX system captures ultrasonic data with a temporal resolution of 0.04 seconds. To ensure robust final predictions, it is necessary to aggregate the classification results over 0.04-second intervals.

During post-processing, the class probabilities generated during inference are arranged in chronological order. The probabilities for all segments within the same 0.04-second interval are aggregated—either by summation or averaging—to compute a representative probability vector for that time slice. For example, if multiple sub-windows exist within a single 0.04-second interval, their probability vectors can be averaged to yield a final probability vector.

If $N$ sliding windows are associated with a specific 0.04-second segment, and the class probability vector for the $i$-th window is $\hat{y}^{(i)} = (\hat{y}_1^{(i)}, \hat{y}_2^{(i)}, ..., \hat{y}_{Class}^{(i)})$, then the representative probability vector $\hat{y}_{final}$ is calculated as follows:

$$\hat{y}_{final} = \frac{1}{N} \sum_{i=1}^{N} \hat{y}^{(i)} \qquad (8)$$

This approach allows the system to determine whether a gas leak or discharge event has occurred within each 0.04-second interval. By aggregating the results, this post-processing step reduces the likelihood of transient false alarms and ensures more reliable classification outcomes.

## 4. Experimental Results

This section presents the results of various experiments conducted to validate the proposed leak and discharge detection system (BATCAM MX) integrated with the Inception-style CNN

Section 4.1 details the dataset, preprocessing steps, and neural network hyperparameters. Section 4.2 describes the defined tasks, while Sections 4.3 to 4.5 compare the proposed method with existing approaches, evaluate classification performance under different signal-to-noise ratio (SNR) conditions, and investigate performance changes in reverberant environments, respectively. Section 4.6 examines the model's generalization capability through k-fold cross-validation, and Section 4.7 discusses model efficiency in terms of parameter count and inference time. Sections 4.8 and 4.9 address, respectively, object detection performance using YOLOv5 and an ablation study that verifies the contribution of the core components in the proposed method.

### 4.1. Dataset and Preprocessing

This study utilizes data recorded by the BATCAM FX acoustic camera[33-35], which consists of 112 microphone channels for beamforming. A band-pass filter is applied to isolate the 20–48 kHz frequency band, effectively excluding lower and audible frequencies. The dataset comprises real gas-leak data recorded in industrial environments, Phase Resolved Partial Discharge (PRPD) signals, and background noise. Figure 6-a presents specimen samples of discharge signals. The discharge dataset includes three types of partial discharge—Corona, Surface, and Floating—collected under various voltage levels by adding 1, 2, and 3 kV above the onset voltage for initial partial discharge specimens. Figure 6-b presents specimen samples of gas leak signals. The gas leak dataset encompasses various piping configurations, such as pinhole, open end, quick connect, and thread coupling, with different leak rates of 200, 500, and 1000 cc/min. All these configurations are labeled as "gas leak signals." Additionally, background noise includes recordings from factory environments and Gaussian noise to simulate diverse industrial scenarios. Therefore, the dataset is composed of a total of five classes: three types of discharge signals, gas-leak signals, and background noise.

Each audio file is sampled at 96 kHz for 10 seconds and converted into a spectrogram using STFT. Each STFT segment processes 512 samples (approximately 0.0053 s) with an overlap of 128 frames, capturing the signal's time–frequency variations. Due to the computational cost of processing the entire spectrogram, a sliding-window technique is applied, dividing the spectrogram into overlapping blocks of 24 frames with an 8-frame hop size. Gamma Correction is then applied to enhance target signal clarity by suppressing irrelevant noise. The preprocessed data are inputted to an 8-layer Inception-style classifier built with TensorFlow 2.x, comprising 19,036 total parameters. The Adam optimizer and categorical crossentropy loss function are used, and a 7:3 split is employed for training and testing. This pipeline ensures robust detection of leaks and discharge signals in complex industrial environments.

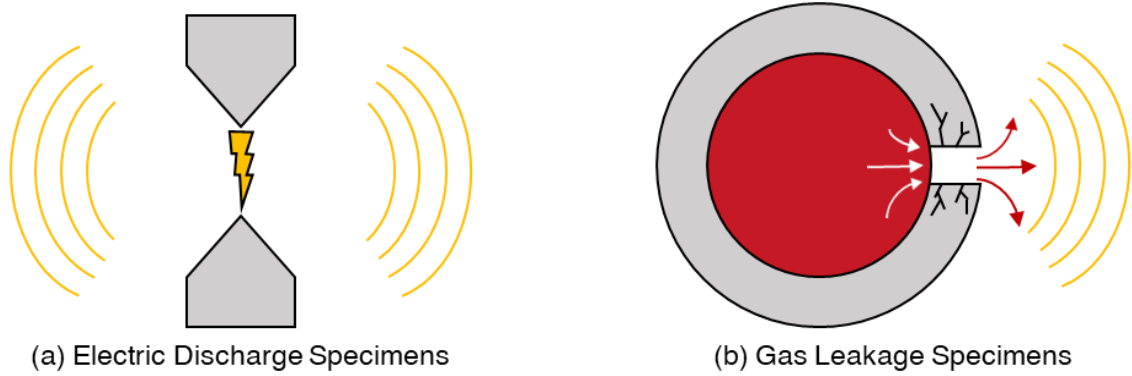

Figure 6. Specimens of gas leakage and PRPD discharge used for training the deep learning model for classification tasks.

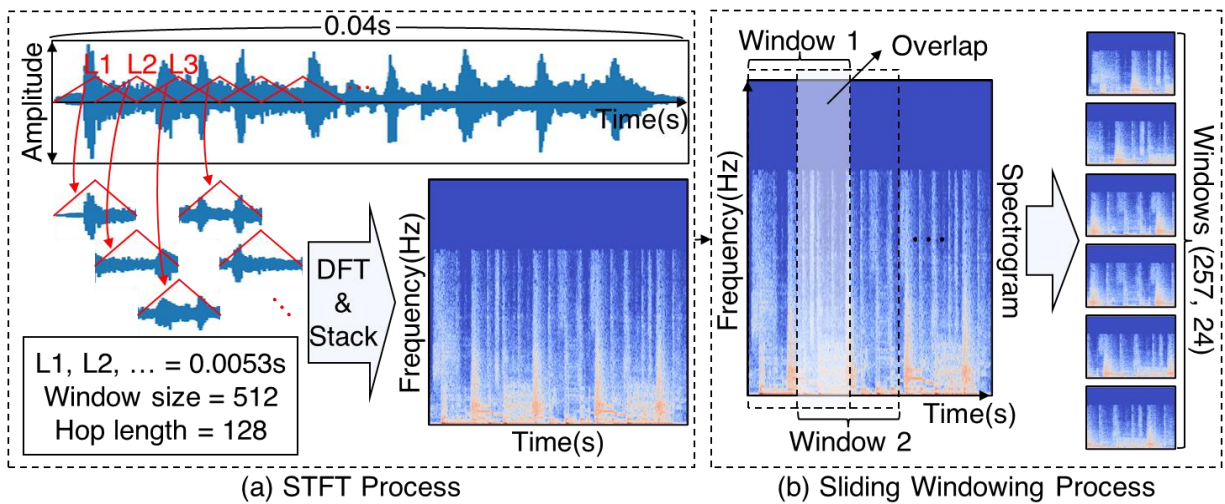

Figure 7. Short-Time Fourier Transform and sliding window process for raw waveform data input to the model. The sliding window is configured with 24 frames and a hop size of 8.

### 4.2. Experiment Setting

As shown in figure 8, the BATCAM MX navigates autonomously within a factory environment to detect gas leakage, partial discharges signals according to four predefined tasks:

- Task A: Detecting gas leak signals in front of the robot.

- Task B: Rotating the Dynamixel motor in the direction specified by a QR code, followed by the detection of PRPD signals.

- Task C: Identifying objects labeled by the YOLOv5 model and determining whether a gas leak is present.
- Task D: Determining the location of specific noise using beamforming and detecting a leak signal in that direction.

These tasks are designed to represent diverse scenarios in industrial settings and to assess the real-world applicability of the proposed system. The performance of the proposed detection and classification system was evaluated using Tasks A, B, C, and D, following these steps:

- k-Fold Cross-Validation: Assess the robustness of the model across the entire dataset.
- YOLOv5 Object Detection Performance: Evaluate the accuracy and recall of target object detection.
- Model Comparison: Compare the performance of the Inception Style CNN with existing deep learning models to validate its superiority.
- Noise Environment Testing: Conduct noise injection experiments to evaluate the model's adaptability to realistic factory noise conditions.
- Spatial Scale Experiments: Assess the adaptability of the deep learning model to varying factory sizes.
- Efficiency Analysis: Measure the parameter size and inference time of the deep learning model to verify its real-time performance.
- Ablation Study: Validate the contributions of the proposed deep learning model's core components.

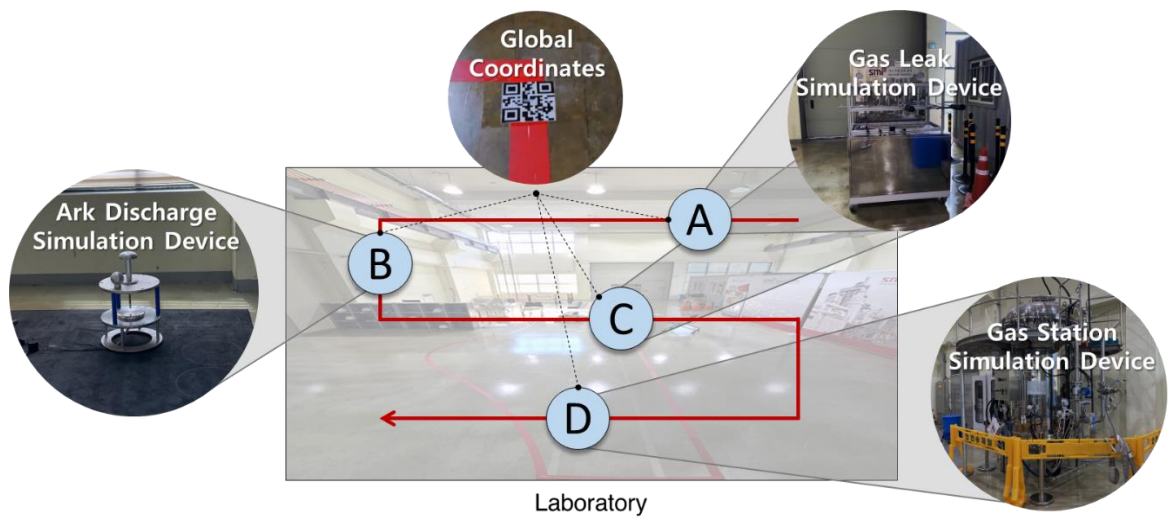

**Figure 8. BATCAM MX autonomously navigates a factory environment, performing tasks such as gas leak and PRPD detection based on QR tags, YOLOv5 object detection, and beamforming.**

Table 2. k-fold cross-validation results showing the precision, recall, F1-score, and accuracy for each fold (k=10). The results indicate consistent model performance and generalizability.

| K-fold | Precision | Recall | F1-Score | Accuracy |
|---|---|---|---|---|
| **1** | 0.92 | 0.95 | 0.92 | 0.92 |
| **2** | 1.00 | 1.00 | 1.00 | 1.00 |
| **3** | 0.97 | 0.99 | 0.98 | 0.98 |
| **4** | 0.95 | 0.95 | 0.95 | 0.93 |
| **5** | 0.99 | 0.99 | 0.99 | 0.99 |
| **6** | 0.99 | 0.99 | 0.99 | 1.00 |
| **7** | 0.99 | 0.99 | 0.99 | 0.99 |
| **8** | 0.97 | 0.98 | 0.97 | 0.98 |
| **9** | 0.89 | 0.92 | 0.89 | 0.89 |
| **10** | 0.88 | 0.92 | 0.89 | 0.91 |

**4.3. k-Fold Validation Test**

To evaluate the robustness and generalization capability of the proposed model, a k-fold cross-validation scheme was adopted. As summarized in Table 2, the entire dataset was divided into 10 subsets, where in each iteration, one subset was used exclusively for testing, and the remaining nine subsets were utilized for training. This process was repeated 10 times, and the average performance across all folds was calculated to ensure balanced evaluation of the model's performance. This validation procedure mitigates potential biases that may arise due to dataset imbalance or specific data distributions, enabling a comprehensive assessment of the model's adaptability to diverse data scenarios.

The results indicate that the variation in F1-scores across folds remained within a narrow margin of 1.2%, demonstrating the robust adaptability of the Inception CNN model to changes in data distribution. This consistent performance across folds underscores the reliability of the proposed model in detecting and classifying signals accurately under varied industrial conditions.

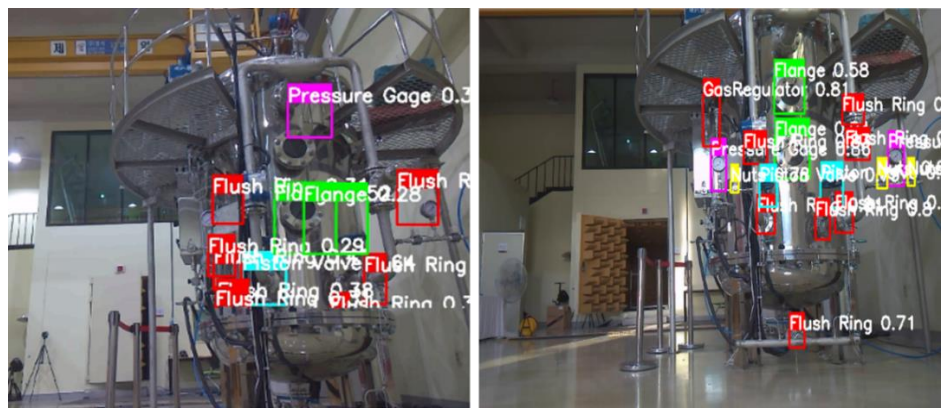

Figure 9. The object detection result captured from BATCAM MX at Task C.
The bounding box displays the predicted label and its probability.

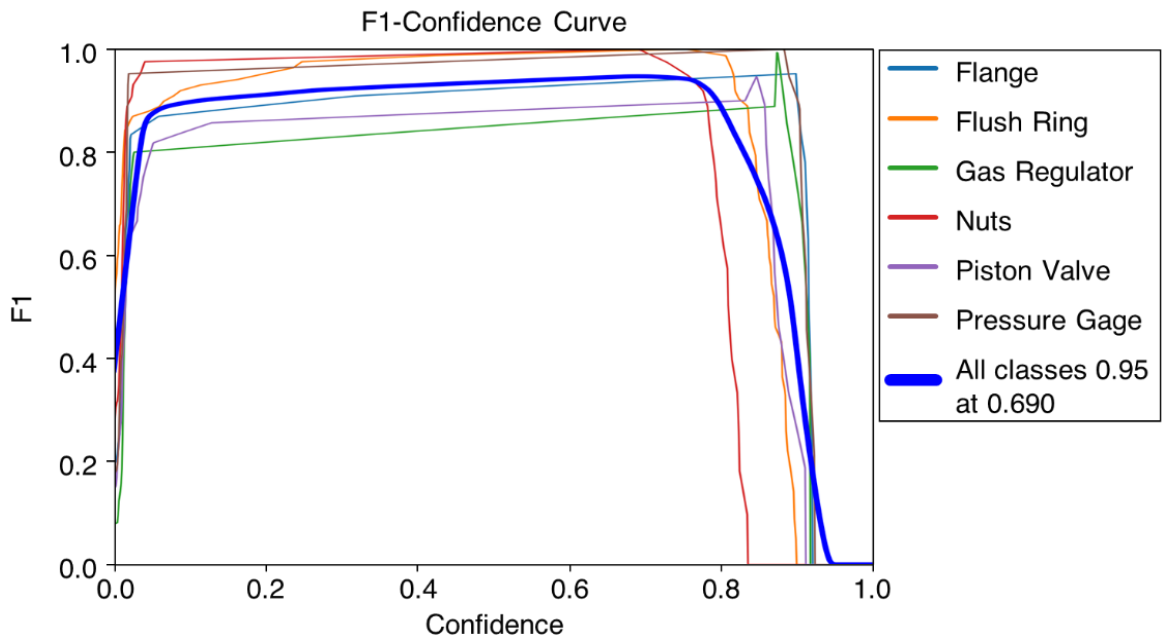

**Figure 10.** F1–Confidence Curve illustrating the object detection performance of YOLOv5 for factory component identification tasks.

### 4.4. Performance Evaluation of YOLOv5-based Video Object Detection

To identify specific components or areas in a factory environment, the BATCAM MX leverages the YOLOv5 object detection model alongside the acoustic detection pipeline. This combination allows for precise localization and classification of objects critical to detecting leak and discharge events.

As illustrated in Figure 10, the F1–confidence curve demonstrates how the F1-score changes across varying confidence thresholds for different factory components. The YOLOv5 model achieves an impressive 97.8% mAP, 94.6% Precision, and 96.8% Recall, indicating robust detection accuracy. Mean Average Precision (mAP) measures the average detection accuracy over a range of Intersection over Union (IoU) thresholds, providing a comprehensive evaluation of the model's ability to locate and classify objects with high precision. The F1-score stabilizes above 0.9 at a confidence threshold of 0.69 for all classes, signifying the model's reliability across diverse object types such as pressure gauges, piston valves, and gas regulators.

These object detection outcomes are integral to guiding the beamforming decisions of the system. For example, after detecting a QR code, the YOLOv5 model identifies and classifies objects within the target area, as shown in Figure 9. The bounding boxes display the predicted labels and their associated probabilities, enabling BATCAM MX to isolate critical components for further acoustic analysis. By integrating the visual outputs from YOLOv5 with the acoustic detection pipeline, the system effectively analyzes leak or discharge signals with high precision. Moreover, the results of the object detection phase inform downstream tasks such as the classification of ultrasonic signals. The integration of visual and acoustic information ensures reliable detection even in complex and noisy factory settings.

**Table 3. Comparison of model performance on raw waveform and spectrogram data using precision, recall, accuracy, and F1-score.**

| Model | | Precision | Recall | F1-Score | Accuracy |
|---|---|---|---|---|---|
| **Raw-Wave Form (1D)** | **CNN** | 0.50 | 0.60 | 0.53 | 0.73 |
| | **LSTM** | 0.50 | 0.60 | 0.53 | 0.73 |
| | **Bi-LSTM** | 0.50 | 0.60 | 0.53 | 0.73 |
| | **MLP** | 0.50 | 0.60 | 0.53 | 0.73 |
| **Spectrogram (2D)** | **CNN** | 0.98 | 0.95 | 0.96 | 0.97 |
| | **Res-Net** | 0.98 | 0.98 | 0.98 | 0.98 |
| | **Ours** | **0.99** | **0.99** | **0.99** | **0.99** |

### 4.5. Comparison with Existing Models

To demonstrate the effectiveness of the proposed model, we conduct a series of comparative experiments against established approaches, as summarized in Table 3. Data representations are categorized into two main types: Raw Waveform and Spectrogram. In the Raw Waveform category, 1D CNN, LSTM, Bi-LSTM, and MLP models are employed, whereas in the Spectrogram category, 2D CNN, ResNet, and the proposed Inception-style Convolutional model are evaluated. These comparisons aim to validate the superior performance of the proposed model in handling complex ultrasonic signals under challenging conditions.

The evaluation is based on four key metrics: Precision, Recall, F1-score, and Accuracy. Precision measures the proportion of correctly identified positive instances among all detected positives, representing the ability to avoid false positives. Recall, on the other hand, quantifies the proportion of true positives identified among all actual positives, reflecting the model's capacity to avoid false negatives. F1-score, calculated as the harmonic mean of Precision and Recall, provides a balanced metric for assessing the overall reliability of the model. Accuracy measures the proportion of correctly classified samples across the entire dataset and offers an aggregate measure of model performance. As detailed in Table 3, models trained on raw waveforms, such as 1D CNN and LSTM-based architectures, exhibit significant performance deterioration in the presence of high levels of noise. These models struggle to extract meaningful features from the waveform domain, which limits their applicability in noisy industrial environments. Conversely, models utilizing Spectrogram-based representations, including 2D CNN and ResNet, demonstrate more stable performance across all metrics. However, the proposed Inception-style CNN consistently achieves the highest scores in Precision, Recall, F1-score, and Accuracy, significantly outperforming its counterparts.

Table 4. F1-score performance comparison of the proposed model and other models under varying SNR(dB) levels with added Gaussian noise.

| SNR (dB) | Raw Waveform(1D) | | | | Spectrogram(2D) | | |
|---|---|---|---|---|---|---|---|
| | CNN | LSTM | Bi-LSTM | MLP | CNN | Res-Net | Ours |
| 5 | 0.54 | 0.54 | 0.54 | 0.34 | 0.86 | 0.83 | **0.97** |
| 4 | 0.48 | 0.53 | 0.50 | 0.29 | 0.73 | 0.79 | **0.94** |
| 3 | 0.33 | 0.50 | 0.45 | 0.21 | 0.58 | 0.75 | **0.90** |
| 2 | 0.23 | 0.43 | 0.28 | 0.16 | 0.45 | 0.74 | **0.84** |
| 1 | 0.21 | 0.41 | 0.22 | 0.14 | 0.40 | 0.71 | **0.73** |
| 0 | 0.17 | 0.40 | 0.20 | 0.13 | 0.38 | **0.59** | 0.58 |
| -1 | 0.12 | 0.40 | 0.16 | 0.12 | 0.38 | **0.51** | 0.43 |
| -2 | 0.11 | 0.41 | 0.12 | 0.11 | 0.37 | **0.46** | 0.26 |
| -3 | 0.11 | **0.43** | 0.11 | 0.11 | 0.33 | 0.30 | 0.13 |

## 4.6. Classification Performance under Varying SNR Levels

The model's classification performance under varying noise levels was evaluated by introducing Gaussian noise ranging from +5 dB to −3 dB into the original ultrasonic signals, as summarized in Table 4. The Signal-to-Noise Ratio (SNR), a key metric in this evaluation, is defined in Equation (9) as the logarithm (in decibels) of the ratio between signal power $P_s$ and noise power $P_n$:

$$SNR = 10 \log_{10}\left(\frac{P_s}{P_n}\right) \quad (9)$$

When the SNR falls below 0 dB, the noise power becomes comparable to or greater than the signal power, significantly complicating the task of extracting target features for most deep learning models. As shown in Table 4, all tested models demonstrate poor performance below −1 dB, reflecting their inability to identify meaningful patterns amidst overwhelming noise. However, performance improves markedly at +1 dB and above, where the signal becomes more distinguishable from the noise.

The proposed model, which utilizes multi-scale filters within its Inception architecture, exhibits superior performance across the +5 dB to +1 dB range. This is attributed to its enhanced ability to capture residual features and isolate critical information masked by noise. Notably, the model maintains the highest F1-score compared to other architectures, demonstrating its robustness in complex noise environments. These findings underscore the effectiveness of the Inception CNN in handling diverse noise conditions, making it a reliable solution for deep learning–based autonomous robotic systems operating in challenging industrial settings.

Table 5. F1-score performance comparison of the proposed model and other models under simulated reverberation using Pyroomacoustics for varying room sizes.

| Model | | Room size($w \times l \times h, m^3$) | | | |
|---|---|---|---|---|---|
| | | $20 \times 10 \times 20$ | $30 \times 15 \times 30$ | $40 \times 20 \times 40$ | $50 \times 25 \times 50$ |
| Raw Waveform | CNN | 0.58 | 0.62 | 0.63 | 0.62 |
| | LSTM | 0.57 | 0.59 | 0.60 | 0.60 |
| | Bi-LSTM | 0.55 | 0.55 | 0.55 | 0.55 |
| | MLP | 0.37 | 0.39 | 0.41 | 0.41 |
| Spectrogram | CNN | 0.48 | 0.53 | 0.60 | 0.71 |
| | Res-Net | 0.82 | 0.99 | 0.98 | 0.98 |
| | **Ours** | **0.92** | **0.99** | **0.99** | **0.99** |

### 4.7. Performance in Reverberant Environments of Different Room Sizes

Large-scale factory buildings introduce overlapping sound reflections, causing reverberation that distorts ultrasonic signals. Reverberation occurs when sound waves reflect off surfaces such as walls, ceilings, and floors, leading to time delays and phase shifts in the original signal. These effects can degrade classification accuracy by reducing the clarity of the target signal. For ultrasonic signals, strong reflections often generate complex distortion patterns in the time–frequency domain, further complicating signal analysis. The interference between direct and reflected signals diminishes the signal-to-noise ratio (SNR) and poses challenges for models attempting to learn critical features accurately. To replicate these conditions, the Pyroomacoustics library is utilized to simulate factory environments of varying sizes: small (30 m²), medium (100 m²), and large (500 m²). Reverberation is closely related to room size and surface reflectivity, with smaller rooms exhibiting shorter reflection paths and more pronounced overlapping interferences. Table 5 demonstrates that smaller rooms introduce more complex reflection interferences, slightly reducing overall classification accuracy.

Nevertheless, the Inception CNN achieves an F1-score approximately 0.14%p higher than ResNet in smaller spaces. This improvement is likely due to the multi-scale filters' ability to effectively capture essential features despite the time delays and phase shifts caused by reverberation. The multi-scale architecture processes interactions across frequency bands, enabling robust signal classification even in challenging reverberant environments. These results highlight the proposed model's reliability and robustness in handling diverse reverberation scenarios, underscoring its suitability for deployment in complex industrial settings.

Table 6. Performance comparison of the proposed model and other models under various conditions in terms of parameter count and inference time.

| Model | | Parameter number | Time(s) | Accuracy | SNR(5dB) F1-Score | Room Size F1-Score (20 × 10 × 20) |
|---|---|---|---|---|---|---|
| Raw Waveform | CNN | 1,382,936 | 8.17 | 0.73 | 0.54 | 0.58 |
| | LSTM | 7,165,445 | 61.17 | 0.73 | 0.54 | 0.57 |
| | Bi-LSTM | 3,558,213 | 109.31 | 0.73 | 0.54 | 0.55 |
| | MLP | 1,836,293 | 0.73 | 0.73 | 0.34 | 0.37 |
| Spectrogram | CNN | **2,973** | 0.97 | 0.97 | 0.86 | 0.48 |
| | Res-Net | 50,077 | **1.1** | 0.98 | 0.83 | 0.82 |
| | **Ours** | 21,810 | 2.1 | **0.99** | **0.97** | **0.92** |

### 4.8. Parameter Count and Performance Comparison

To assess real-time feasibility, each model's total parameters and inference time are measured. As reported in Table 6, the Inception CNN comprises 19,036 parameters—relatively few—yet achieves inference within about 2.1 s and maintains top-tier performance in the SNR and reverberation experiments. Although ResNet exhibits shorter inference times, it generally requires more parameters and shows somewhat inferior performance in practical scenarios involving noisy or reverberant signals. Overall, the proposed approach offers a well-balanced trade-off among accuracy, parameter efficiency, and inference speed.

### 4.9. Ablation Study

As shown in Table 7, an ablation study was conducted to measure how the Inception architecture and Gamma Correction affect performance. As presented in, removing both elements' results in a noticeable performance decline, with the F1-score dropping by up to 0.14%p. Notably, removing the Inception module causes a more pronounced degradation than removing Gamma Correction, indicating that multi-scale feature extraction is pivotal for sustaining stable performance. Nonetheless, Gamma Correction also plays a vital role by intensifying relevant regions in the beamformed target signal while attenuating background noise. Hence, both components operate synergistically to maximize detection accuracy for leaks and partial discharges across diverse industrial environments.

By synthesizing these experimental findings, the proposed Inception-style CNN, combined with beamforming and YOLOv5-based object recognition, demonstrates high reliability in detecting ultrasonic anomalies—such as gas leaks and partial discharges—in industrial settings characterized by complex noise and reverberation. The integration of QR code navigation further ensures real-time feasibility for autonomous vehicles operating on factory floors.

Table 7. Ablation study results: performance impact of removing the Inception module and Gamma Correction on detection metrics.

|  | Precision | Recall | F1-Score | Accuracy |
|---|---|---|---|---|
| **w/o Inception, Gamma** | 0.94 | 0.83 | 0.82 | 0.89 |
| **w/o Gamma** | 0.96 | 0.91 | 0.92 | 0.94 |
| **w/o Inception** | 0.95 | 0.85 | 0.86 | 0.90 |
| **Ours** | **0.98** | **0.95** | **0.96** | **0.97** |

## 5. Concluding Remarks

This study proposed the BATCAM MX, a deep learning–based autonomous robotic system designed for the detection and classification of gas leaks and arc discharges in industrial environments. Inspired by the Human Protocol, the BATCAM MX integrates visual information to identify targets and autonomously navigates to optimal positions for collecting and analyzing acoustic data. The system leverages the BATCAM FX's 112-channel microphone array, capable of capturing ultrasonic signals up to 48 kHz. These signals are refined through a beamforming and STFT–based acoustic processing pipeline. By employing Gamma Correction and an Inception-style CNN, the system extracts features across multiple time–frequency scales, achieving a high detection accuracy of 99%.

Notably, the proposed system demonstrated exceptional performance in challenging environments characterized by reverberation and intense noise. Across four controlled experimental tasks (Tasks A, B, C, D), the BATCAM MX outperformed conventional approaches, achieving up to a 44 percentage-point improvement in performance. Validation with real-world manufacturing scenarios further highlighted the practical applicability of the system. This emphasizes the robustness and effectiveness of integrating visual and acoustic information, modeled after the Human Protocol, as a practical solution for industrial automation.

Additionally, the system maintained an inference time of 2.1 seconds, showcasing its ability to systematically execute highly complex operations through the interaction of autonomous navigation and deep learning models. These capabilities underline the system's potential to significantly enhance safety and operational efficiency in industrial settings.

Future research will focus on improving signal separation under extreme noise conditions ($\leq 0$ dB) by exploring advanced acoustic noise suppression techniques specifically tailored to ultrasonic frequency ranges. To achieve this, lightweight CNN-based models will be developed to efficiently suppress noise while preserving critical ultrasonic signals [36].

## Acknowledgments

This research was supported by SM Instruments Inc.